\documentclass[conference,a4paper]{IEEEtran}
\IEEEoverridecommandlockouts

\usepackage[hidelinks]{hyperref}
\usepackage[cmex10]{amsmath}
\usepackage{amssymb,amsfonts}
\usepackage{dblfloatfix}

\usepackage[ruled,vlined]{algorithm2e}
\usepackage{graphicx}
\graphicspath{{Figures/PDF/}{Figures/PNG/}}
\usepackage{rotating}
\usepackage{subcaption} 
\usepackage{booktabs}
\usepackage{siunitx}
\usepackage[numbers,compress]{natbib}
\usepackage{texnames}
\usepackage{bm,bbm}
\usepackage{orcidlink}
\usepackage{xcolor}
\usepackage{soul}
\usepackage{float}
\begin{document}


\title{\uppercase{Efficient Adaptation For Remote Sensing Visual Grounding}
}


\author{	
    \IEEEauthorblockN{Hasan Moughnieh}
	\IEEEauthorblockA{\textit{American University of Beirut}\\
		Beirut , Lebanon}
        \and
        
	\IEEEauthorblockN{Mohamad Chalhoub}
	\IEEEauthorblockA{\textit{Lebanese University}\\
		Beirut , Lebanon}
	\and
	
    \IEEEauthorblockN{Hasan Nasrallah}
	\IEEEauthorblockA{\textit{RASID SARL}\\
		Beirut , Lebanon}
    
	\and
	\IEEEauthorblockN{Cristiano Nattero}
	\IEEEauthorblockA{\textit{WASDI}\\
		Dudelange, Luxembourg}
	\and
    
	\IEEEauthorblockN{Paolo Campanella}
	\IEEEauthorblockA{\textit{WASDI}\\
		Dudelange, Luxembourg}

    \and

    \IEEEauthorblockN{Giovanni Nico}
	\IEEEauthorblockA{\textit{Institute for Applied Mathematics, CNR}\\
		Bari, Italy}

    \and
    
    \IEEEauthorblockN{Ali J. Ghandour*\thanks{*Correspondence: aghandour@cnrs.edu.lb.}}
	\IEEEauthorblockA{\textit{National Center for Remote Sensing, CNRS}\\
		Beirut , Lebanon}
}

\maketitle
\begin{abstract}

Adapting pre-trained models has become an effective strategy in artificial intelligence, offering a scalable and efficient alternative to training models from scratch. In the context of remote sensing (RS), where visual grounding(VG) remains underexplored, this approach enables the deployment of powerful vision-language models to achieve robust cross-modal understanding while significantly reducing computational overhead.

To address this, we applied Parameter Efficient Fine Tuning (PEFT) techniques to adapt these models for RS-specific VG tasks. Specifically, we evaluated LoRA placement across different modules in Grounding DINO and used BitFit and adapters to fine-tune the OFA foundation model pre-trained on general-purpose VG datasets. This approach achieved performance comparable to or surpassing current State Of The Art (SOTA) models while significantly reducing computational costs.

This study highlights the potential of PEFT techniques to advance efficient and precise multi-modal analysis in RS, offering a practical and cost-effective alternative to full model training.
    
\end{abstract}


\section{Introduction}

Visual grounding(VG) \cite{yu2018mattnet,wu2022referring,yang2022lavt}, a process designed to identify and localize objects or regions within an image based on natural language descriptions, has advanced significantly; however, their application to the remote sensing (RS) domain remains underexplored. These models struggle when dealing with RS imagery due to its rich contextual information and the variety of spatial resolutions.

VG offers significant potential for RS by allowing seamless interaction with complex geospatial imagery. It enables tasks such as identifying buildings or ships from text descriptions and pinpointing areas affected by disasters or environmental changes. By addressing challenges such as handling large datasets and improving object identification, visual grounding improves the efficiency and accuracy of applications such as urban planning, disaster response, and surveillance.

Foundation models\cite{alayrac2022flamingo,radford2021learningtransferablevisualmodels,li2021align,li2022blip,jia2021scaling}, have changed the field of artificial intelligence by demonstrating remarkable adaptability in different areas. These models, pre-trained on vast datasets, excel in multi-modal tasks by integrating diverse modalities such as textual, visual, audio, video, and sensory data. Their ability to generalize across diverse applications has solidified their role as a foundation for the advancement of AI research and practical implementations. Although models such as Grounding DINO\cite{liu2024groundingdinomarryingdino} and One-For-ALL(OFA)\cite{wang2022ofaunifyingarchitecturestasks} have shown promise in general multi-modality and more specifically visual grounding tasks, their direct application to remote sensing (RS) remains limited. This paper focuses on adapting these powerful models to the unique challenges and data characteristics of RS, enabling their effective use for satellite and aerial imagery analysis.

This raises a critical question: How can general visual grounding models be effectively adapted to address the unique characteristics of RS imagery while maintaining computational efficiency and achieving SOTA performance?


Visual Language Models (VLMs)\cite{xiao2023florence2advancingunifiedrepresentation,li2022blipbootstrappinglanguageimagepretraining,lu2024deepseek,gemini2024family,liu2023improved,wu2024visionllm}, face challenges in RS due to the distinct patterns and complex spatial relationships in RS imagery. Addressing these challenges has been the focus of several SOTA approaches, which aim to bridge the gap between textual descriptions and RS imagery. GeoVG\cite{10.1145/3503161.3548316} introduced a fusion module that integrates geospatial relation graphs and adaptive region attention to align textual queries with RS imagery. The Multi-Granularity Visual Language Fusion (MGVLF)\cite{10056343} module enhances RSVG by leveraging multi-scale features and adaptive noise filtering for better localization. LPVA\cite{10584552} innovates with a Progressive Attention (PA) module to dynamically adjust visual features at the spatial and channel levels, and a Multilevel Feature Enhancement (MFE) decoder to aggregate contextual information and suppress background noise, allowing precise object identification in RS VG tasks. However, these models are generally trained from scratch, requiring extensive computational resources and making their deployment more challenging for broader applications.

This study explores the application of Parameter Efficient Fine Tuning (PEFT) techniques to adapt Grounding DINO, a visual grounding model, and OFA, a foundational vision-language model (VLM), for remote sensing tasks. Using the DIOR-RSVG\cite{10056343} and OPT-RSVG\cite{10584552} datasets, we evaluate the effectiveness of these techniques in addressing the computational challenges and domain-specific requirements of remote sensing, aiming to achieve competitive performance with reduced resource demands.

In this study, we make the following key contributions:
\begin{itemize}
  \item We applied the LoRA PEFT technique to Grounding DINO, achieving a best-performing configuration that outperformed SOTA on the DIOR-RSVG and OPT-RSVG datasets.
\item We compared adapter and bitFit PEFT techniques on OFA, with adapter showing competitive results against Visual Grounding SOTA on the DIOR-RSVG dataset.
\item Using adapter, the best-performing technique on OFA, we trained and evaluated the model on the OPT-RSVG dataset, validating its effectiveness.
\end{itemize}

\section{Methodology}
\subsection{PEFT}
In this research, we leverage Parameter-Efficient Fine Tuning (PEFT)\cite{han2024parameterefficientfinetuninglargemodels} to adapt vision-language models (VLMs) for remote sensing visual grounding. PEFT fine-tunes only a minimal set of parameters, offering computational efficiency while retaining the strengths of pre-trained models.

\subsubsection{Adapter}
Adapters\cite{houlsby2019parameterefficienttransferlearningnlp} are lightweight modules inserted between the layers of pre-trained models. They introduce a small number of additional parameters for task-specific learning while keeping most weights frozen. This makes adapters computationally efficient and effective for remote sensing visual grounding tasks, enabling the model to capture task-specific knowledge such as object localization and scale variations.

\subsubsection{Low-Rank Adaptation (LoRA)}
LoRA\cite{hu2021loralowrankadaptationlarge} reduces trainable parameters by applying low-rank updates to the model weight matrices:
\[
W' = W + \Delta W, \quad \Delta W = A \cdot B,
\]
where $A \in \mathbb{R}^{d \times r}$, $B \in \mathbb{R}^{r \times d}$, and $r \ll d$. Here, $W'$ is the adapted weight matrix, $W$ is the pre-trained matrix, and $\Delta W$ is the low-rank update. LoRA preserves original weights while learning task-specific updates, significantly reducing memory and computation requirements, making it ideal for large-scale visual grounding tasks.

\subsubsection{BitFit}
BitFit\cite{zaken2022bitfitsimpleparameterefficientfinetuning} fine-tunes only the bias terms of the pre-trained model's layers:
\[
W' = W + \Delta b,
\]
where $\Delta b$ represents task-specific bias adjustments. By updating only the bias terms, BitFit minimizes computational overhead while preserving the majority of pre-trained parameters, making it a highly efficient method for adapting models to specific tasks.

\subsection{Remote Sensing Visual Grounding datasets}
Datasets play a critical role in advancing Visual Grounding in RS, yet their availability remains limited. RSVG-H\cite{10.1145/3503161.3548316} was the first dataset designed for this task, consisting of 4,239 RS images, 7,933 text expressions, and 5,994 objects. Although the data set captures complex geospatial relationships, its limited size and the difficulty of locating a precise target constrain its applicability. DIOR-RSVG\cite{10056343} expanded on this with 17,402 images and 38,320 language expressions spanning 20 categories. However, the class imbalance in DIOR-RSVG impacts model performance, often biasing predictions toward dominant classes.
To address this, the OPT-RSVG\cite{10584552} dataset introduced 25,452 RS images and 48,952 image-query pairs in 14 object categories, incorporating both English and Chinese annotations for broader adaptability. In this paper, we will use both the DIOR-RSVG and OPT-RSVG datasets.

\subsection{Evaluation Metrics}

A predicted bounding box for a remote sensing (RS) image-query pair is correct if the Intersection-over-Union (IoU) with the ground-truth box exceeds a threshold, with 0.5 commonly used as a baseline accuracy metric. We report precision metrics at IoU thresholds of 0.5, 0.7, and 0.9, denoted as Pr@0.5, Pr@0.7, and Pr@0.9.

Model performance is further evaluated using mean IoU and cumulative IoU:
\begin{equation}
\text{meanIoU} = \frac{1}{M} \sum_{t} \frac{I_t}{U_t},
\end{equation}
\begin{equation}
\text{cumIoU} = \frac{\sum_{t} I_t}{\sum_{t} U_t},
\end{equation}
where $t$ indexes the image-query pairs, $M$ is the total number of pairs, and $I_t$, $U_t$ are the intersection and union areas.

To assess computational efficiency, we calculate the proportion of frozen parameters:
\begin{equation}
\text{Efficiency} = 100 - \left( \frac{\text{Trainable Parameters (PEFT)}}{\text{Total Model Parameters}} \times 100 \right)
\end{equation}

This metric highlights the reduced computational cost achieved through PEFT techniques by reflecting the percentage of the model that remains static during training.

\subsection{Model Selection}
This study uses two SOTA models: OFA and Grounding DINO, chosen for their demonstrated efficacy in vision-language tasks and VG. OFA serves as a foundational model, offering a robust pre-trained architecture designed for multi-modal tasks. Its unified image-text capabilities make it highly versatile across diverse applications. Grounding DINO, on the other hand, is specifically tailored for visual grounding, leveraging cross-modal attention to accurately associate text with visual regions. Its integration into cutting-edge frameworks such as the Text Segment Anything Model (SAM)\cite{kirillov2023segment} underscores its versatility and effectiveness in VG tasks, positioning it as an ideal candidate for addressing remote sensing challenges.

\section{Experimental Results}
\subsection{Grounding DINO}
Grounding DINO is a SOTA VG model that takes advantage of a transformer-based architecture with cross-modal attention to link text prompts to specific regions in an image. It utilizes a vision transformer (ViT) backbone to extract rich image features and a text encoder to process input prompts. Multi-scale deformable attention further enhances the model's ability to integrate features from various resolutions, enabling accurate object detection and phrase localization. By adopting a unique query design, Grounding DINO achieves top-tier performance in tasks such as object detection, visual grounding, and phrase localization.

In this work, Grounding DINO is divided into three distinct modules: the text encoder, the image encoder, and the decoder with integrated cross-modal attention. To enable parameter-efficient fine-tuning, we incorporated LoRA layers with a rank of 16 into all dense (fully connected) layers across these modules. This modification allows the model to adapt to the specific requirements of VG tasks while maintaining the majority of its pre-trained parameters intact. The experiments were structured to evaluate the effect of integrating LoRA into individual modules and combinations of modules. These configurations were designed to identify the optimal balance between computational efficiency and task performance for VG. The variants were trained using the DIOR-RSVG training set and their performance was evaluated on its test set. The results of this evaluation are presented in Table \ref{lora placement}.

Finally, the optimal LoRA placement configuration was evaluated against full fine-tuning and SOTA models. The comparative results, which demonstrate its efficiency and effectiveness, are presented in Table \ref{SOTA comparison}.

\begin{table}[ht]
\caption{Performance metrics for different LoRA-based PEFT configurations applied to the Grounding DINO model evaluated on the DIOR-RSVG test set.  All numbers are percentages (\%).}
\centering
\scriptsize 
\setlength{\tabcolsep}{2pt} 
\renewcommand{\arraystretch}{1.3} 
\begin{tabular}{l|c|ccc|c}
\hline
\textbf{Methods}              & \textbf{Efficiency} & \textbf{Pr@0.5} & \textbf{Pr@0.7} & \textbf{Pr@0.9} & \textbf{meanIoU} \\ \hline
Image Encoder                 & 98.66                       & 54.1            & 47.3            & 27.8            & 48.8                   \\
Decoder                       & 99.05                       & 78.1            & 71.5            & 43.2            & 70.0                   \\
Image Encoder + Decoder       & 97.70                       & 81.1            & 74.1            & 44.3            & 82.8                   \\
\textbf{Encoders + Decoder}   & \textbf{96.74}              & \textbf{81.3}   & \textbf{74.7}   & \textbf{45.2}   & \textbf{82.9}          \\ \hline
\end{tabular}
\vspace{2pt}
\label{lora placement}
\end{table}

\subsection{One For ALL (OFA)}
The OFA\cite{wang2022ofaunifyingarchitecturestasks} model is a transformer-based framework for multi modal tasks like image captioning, visual question answering, and visual grounding. Its encoder-decoder architecture processes visual patches and textual tokens, learning rich joint representations through cross-attention for precise alignment. Pre-trained on large multi modal datasets, OFA generalizes well across tasks while allowing efficient fine-tuning for specialized applications, such as RS VG. Its unified design reduces task-specific overhead, making it a versatile and effective model for VG tasks.

To adapt the OFA model for Remote Sensing VG tasks, we employed a PEFT strategy that incorporates adapters into its architecture. Most of the pre-trained parameters of the model, such as embeddings, attention mechanisms, and fully connected layers, were kept frozen to retain the extensive knowledge gained during pre-training. Task-specific learning was facilitated by introducing trainable adapter layers, specifically down-projection and up-projection modules, into both the encoder and decoder components. This integration effectively captures domain-specific features while significantly minimizing the number of trainable parameters, ensuring computational efficiency without compromising model performance.

The BitFit technique employs selective fine-tuning by freezing almost all parameters of the model, with the exception of biases. For the OFA model, the vast majority of components, including the embedding layers, convolutions, and attention mechanisms in both the encoder and decoder, were frozen. Only a small subset of parameters, mainly the biases in layer normalization, fully connected layers, and certain attention layers, were trainable. This approach reduced the number of trainable parameters to just 70,656 out of the model's total 182 million, showcasing BitFit's effectiveness in resource-efficient fine-tuning for domain-specific tasks.

The results of these experiments, comparing the efficiency and performance of BitFit and Adapters in fine-tuning the OFA model on the DIOR-RSVG test set with SOTA models, are presented in Table \ref{SOTA comparison}.

\begin{table}[ht]
\caption{Comparison of SOTA visual grounding models and our proposed PEFT methods. Grounding DINO with LoRA and OFA with Adapter and Bitfit on the DIOR-RSVG test set. All numbers are percentages (\%).}

\centering
\setlength{\tabcolsep}{1pt}
\renewcommand{\arraystretch}{1} 
\begin{tabular}{l|ccc|c|c}
\hline
\textbf{Methods}              & \textbf{Pr@0.5} & \textbf{Pr@0.7} & \textbf{Pr@0.9} & \textbf{meanIoU} & \textbf{cumIoU} \\ \hline
TransVG\cite{deng2022transvgendtoendvisualgrounding}                       & 72.41           & 60.05           & 27.84           & 63.56           &   76.27               \\
VLTVG (ResNet-50) \cite{9879831}            & 69.41           & 58.44           & 24.37           & 59.96           &  71.97                \\
VLTVG (ResNet-101)            & 75.79           & 66.33           & 33.11           & 66.32           &      77.85            \\
QRNet\cite{ye2022shiftingattentionvisualbackbone}                          & 75.84          & 62.27           & 25.69           & 66.80           &       75.39           \\
MGVLF\cite{10056343}                          & 76.78           & 66.74           & 35.07           & 68.04           & 78.41                 \\ 
LPVA\cite{10584552}                       & \textbf{82.27}  & 72.25           & 39.55           & 72.35           &  \textbf{85.11}                \\ \hline
Grounding DINO (Vanilla)       & 26.6            & 20.1            & 8.8             & 28.1            &        20.0          \\ 
Grounding DINO (FFT)             & 76.8            & 68.4            & 38.1           & 67.5            &     76.3             \\ \hline
GroundingDINO+LoRA (Ours)   & 81.3   & \textbf{74.7}   & \textbf{45.2}  & \textbf{82.9}   &         80.1         \\ \hline
OFA + Adapter (Ours)     &  76.72            & 63.14            & 30.07            & 62.23   &   77.33       \\ 
OFA + BitFit (Ours)  & 56.97           & 44.22           & 18.95           & 37.7    & 40.2          \\ \hline
\end{tabular}
\vspace{1pt}
\label{SOTA comparison}
FFT\textsuperscript{†} denotes: Full Fine-Tuning
\end{table}
To evaluate the generalization capabilities of PEFT techniques, we conducted experiments on the English subset of the OPT-RSVG dataset. In contrast to DIOR-RSVG, which contains 20 object categories, OPT-RSVG presents unique challenges with its 14 categories, increased dataset size of 25,452 images of diverse resolutions, and greater linguistic complexity in textual expressions, averaging 10.10 words compared to DIOR-RSVG's 7.47. For Grounding DINO, we employed the DIOR-RSVG pre-trained image encoder and decoder with a LoRA configuration, which achieved a strong balance between efficiency and performance, as demonstrated in \ref{lora placement}. Similarly, we utilized the DIOR-RSVG pre-trained OFA model with adapter modules. Both models were fine-tuned on the English subset of the OPT-RSVG dataset. 
These experiments underscore the adaptability and computational efficiency of PEFT methods in addressing the challenges of remote sensing visual grounding datasets with diverse characteristics and intricate textual descriptions. The comparative results with SOTA models are presented in Table \ref{tab:opt-rsvg-sota}.

\begin{table}[ht]
\caption{Comparison with the SOTA Methods for Adapted OFA and Grounding DINO on the Test Set of OPT-RSVG (English Version). All numbers are percentages (\%) }
\centering
\setlength{\tabcolsep}{2pt}
\renewcommand{\arraystretch}{1} 
\begin{tabular}{l|ccc|c|c}
\hline
\textbf{Methods} & \textbf{Pr@0.5} & \textbf{Pr@0.7} & \textbf{Pr@0.9} & \textbf{meanIoU} & \textbf{cumIoU} \\ \hline
TransVG\cite{deng2022transvgendtoendvisualgrounding}  & 69.96 & 54.68 & 12.75 & 59.80 & 69.31 \\
VLTVG (ResNet-50)\cite{9879831}  & 71.84 & 57.79 & 14.53 & 61.44 & 70.69 \\
VLTVG (ResNet-101)  & 73.50 & 63.11 & 16.31 & 62.48 & 73.86 \\
MGVLF\cite{10056343}  & 72.19 & 58.86 & 15.10 & 61.51 & 71.80 \\
LPVA\cite{10584552}  & 74.69 & 60.56 & 15.84 & 63.78 & \textbf{74.42} \\ \hline
OFA+Adapter (Ours) & 66.38 & 46.7 & 12.86 & 41.67 & 66.39 \\\hline
Grounding DINO (Ours) & \textbf{75.81} & \textbf{66.47} & \textbf{26.39} & \textbf{65.24} & 69.53 \\\hline
\end{tabular}
\label{tab:opt-rsvg-sota}
\end{table}
\vspace{0.25pt}

\section{Discussion}

The results detailed in Tables \ref{lora placement}, \ref{SOTA comparison}, and \ref{tab:opt-rsvg-sota} provide an in-depth analysis of how parameter-efficient fine-tuning (PEFT) techniques were applied to adapt both the Grounding DINO and OFA models for VG tasks in RS. These results emphasize the balance between computational efficiency and performance, showcasing the effectiveness of PEFT techniques.

\begin{figure}[H]
    \centering
    \tiny
    \textbf{DIOR-RSVG Dataset}\par\vspace{2pt}
    \begin{tabular}{c c}
        \begin{minipage}[t]{0.15\textwidth}
            \centering
            \includegraphics[width=\linewidth]{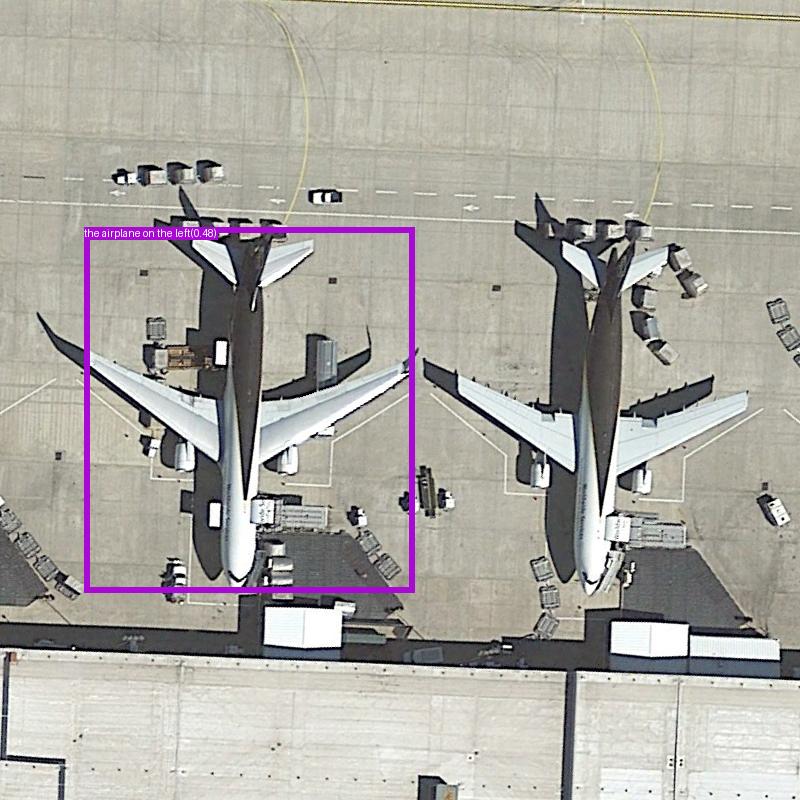}
            {\footnotesize{The airplane on the left}}
        \end{minipage}
        &
        \begin{minipage}[t]{0.15\textwidth}
            \centering
            \includegraphics[width=\linewidth]{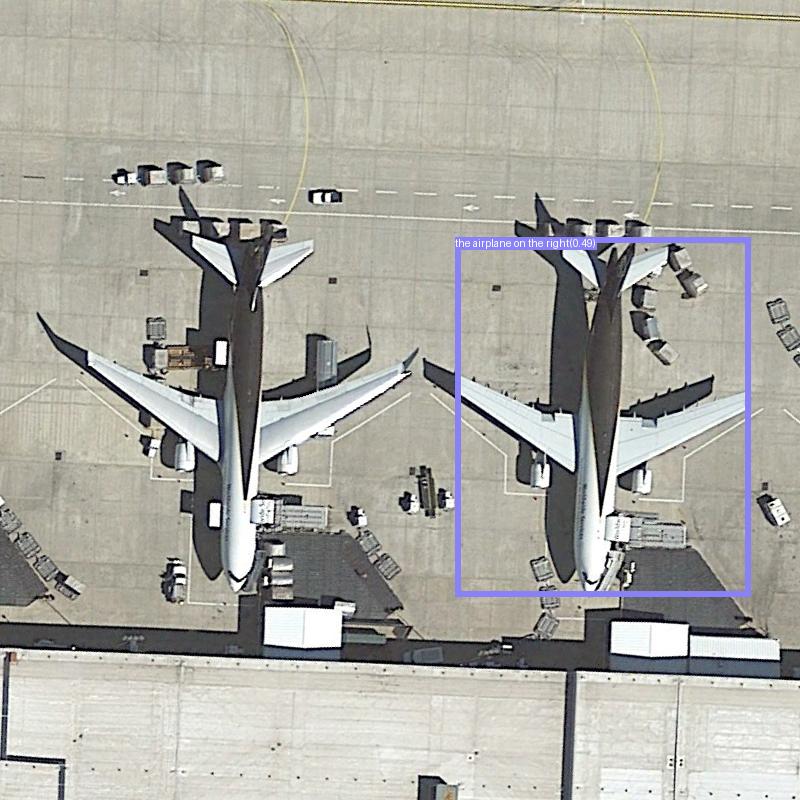}
            {\footnotesize{The airplane on the right}}
        \end{minipage}
        \\[1pt]
        \begin{minipage}[t]{0.15\textwidth}
            \centering
            \includegraphics[width=\linewidth]{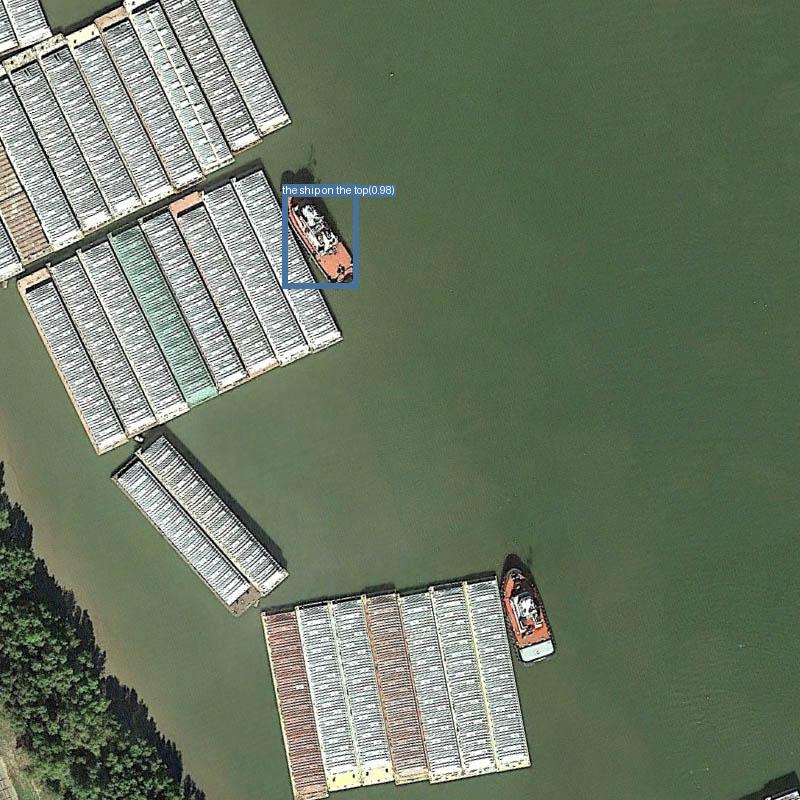}
            {\footnotesize{The ship on the top}}
        \end{minipage}
        &
        \begin{minipage}[t]{0.15\textwidth}
            \centering
            \includegraphics[width=\linewidth]{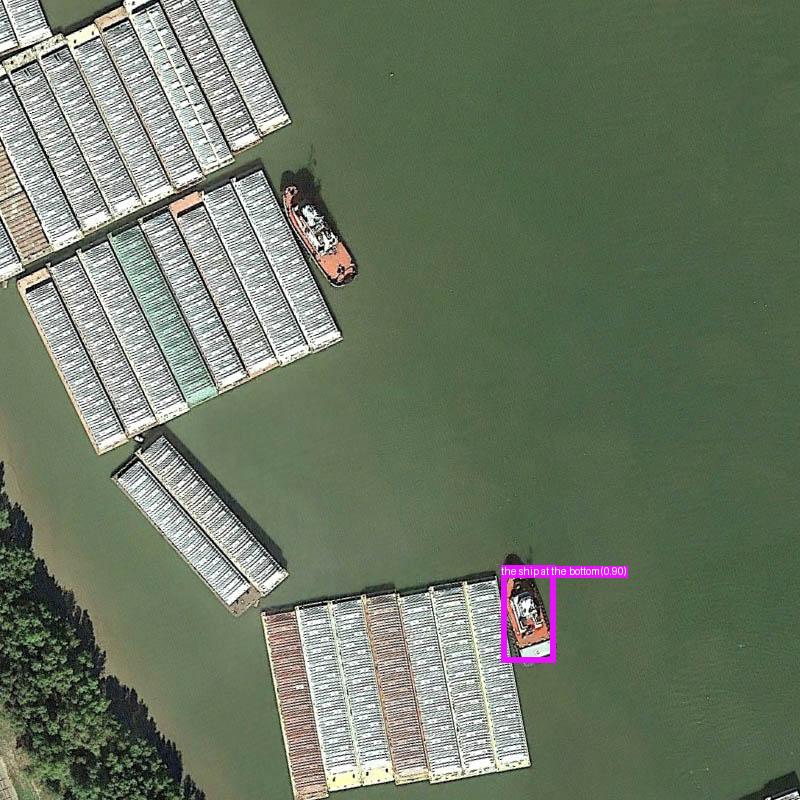}
            {\footnotesize{The ship at the bottom}}
        \end{minipage}
    \end{tabular}

    \vspace{5pt}
    \textbf{OPT-RSVG Dataset}\par\vspace{4pt}
    \begin{tabular}{c c}
        \begin{minipage}[t]{0.15\textwidth}
            \centering
            \includegraphics[width=\linewidth]{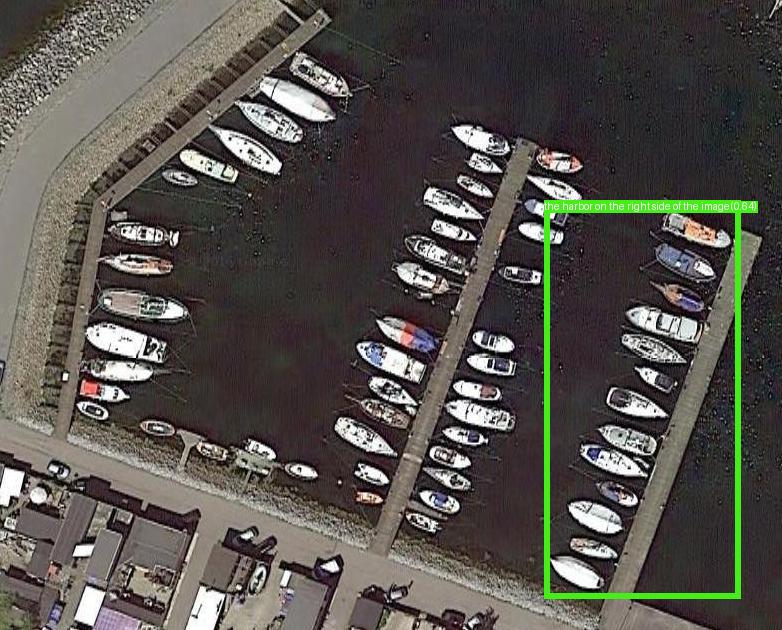}
            {\footnotesize{The harbor on the right side}}
        \end{minipage}
        &
        \begin{minipage}[t]{0.15\textwidth}
            \centering
            \includegraphics[width=\linewidth]{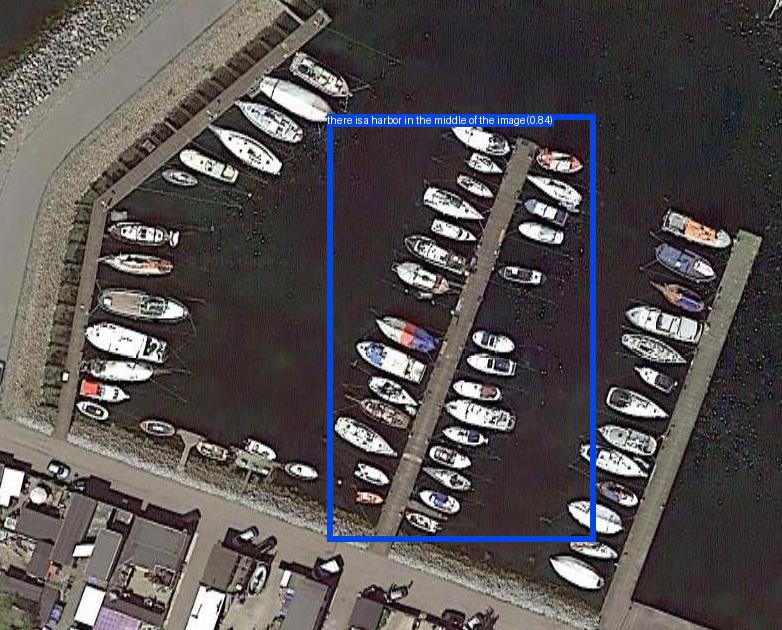}
            {\footnotesize{There is a harbor in the middle}}
        \end{minipage}
        \\[5pt]
        \begin{minipage}[t]{0.15\textwidth}
            \centering
            \includegraphics[width=\linewidth]{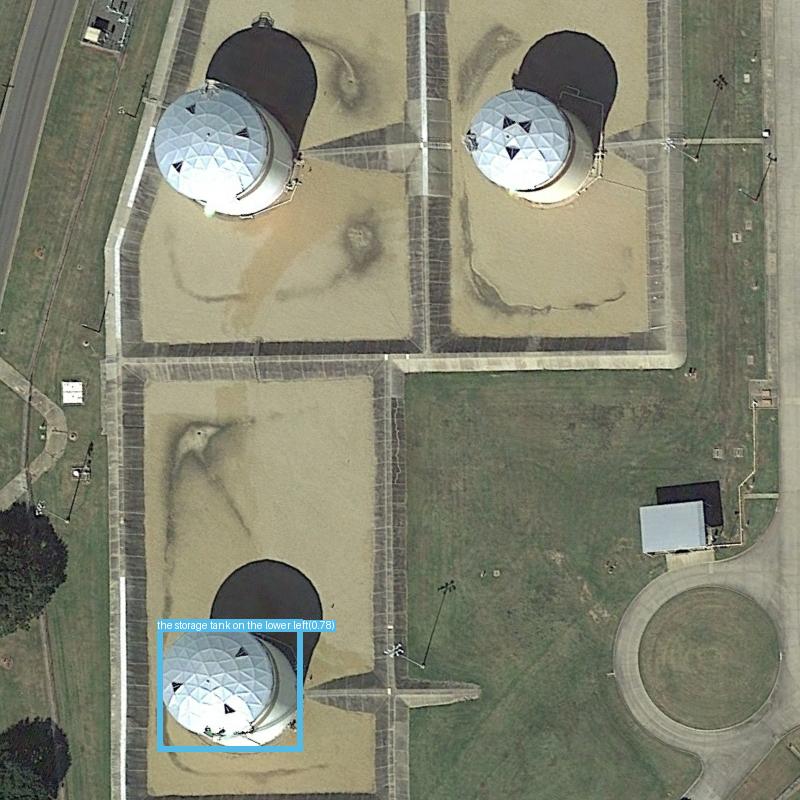}
            {\footnotesize{The storage tank on the lower left}}
        \end{minipage}
        &
        \begin{minipage}[t]{0.15\textwidth}
            \centering
            \includegraphics[width=\linewidth]{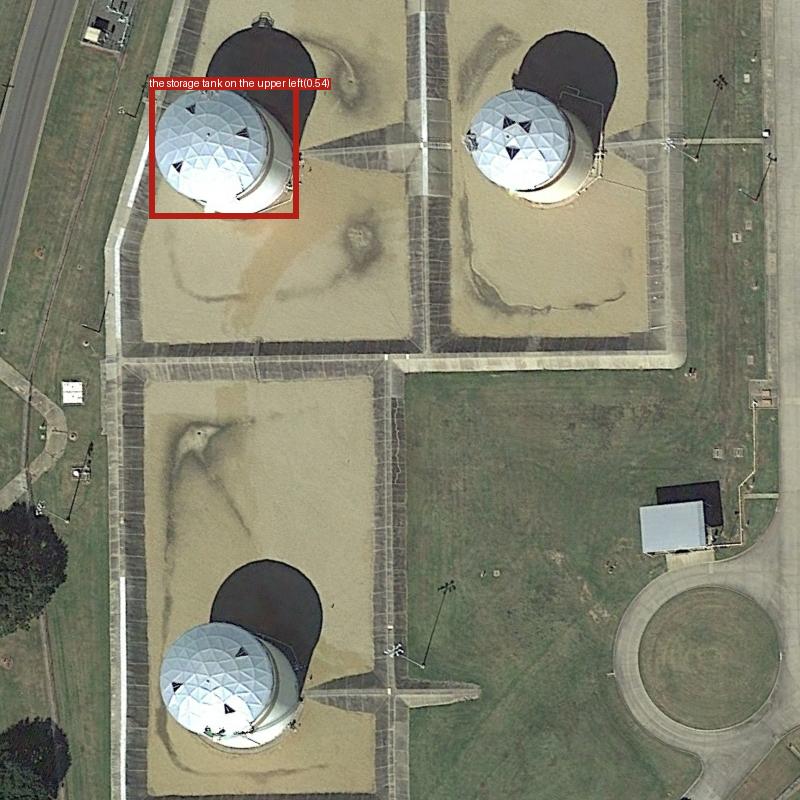}
            {\footnotesize{The storage tank on the upper left}}
        \end{minipage}
    \end{tabular}

    \caption{LoRA-adapted Grounding DINO inference results on DIOR-RSVG and OPT-RSVG datasets.}
    \label{fig:inference}
\end{figure}

The performance of Grounding DINO with various LoRA placements, as shown in \ref{lora placement}, reveals interesting insights. The configuration involving both the image and decoder modules ("Encoders + Decoder") achieves the highest meanIoU of 82.9\% and a Pr@0.9 of 45.2\%. Notably, adding the text encoder to the LoRA modules minimally contributed to performance gains, as demonstrated by the slight improvement over the "Image Encoder + Decoder" setup. This finding highlights that focusing LoRA adjustments on the image encoder and decoder is sufficient to achieve high performance. Furthermore, this setup trained only 3.26\% of the total parameters, underscoring its computational efficiency. Compared to full fine-tuning, which updates all parameters , LoRA achieves better results while dramatically reducing resource requirements. As shown in Table \ref{SOTA comparison}, LoRA outperformed previous state-of-the-art visual grounding models on the DIOR-RSVG dataset, achieving the highest meanIoU and competitive precision metrics. To further demonstrate the adaptability of our approach, we fine-tuned the LoRA adapted Grounding DINO on the more challenging OPT-RSVG dataset and evaluated it on its corresponding test set. Despite the complexity of the dataset, our method exceeded state-of-the-art baselines in almost all evaluation metrics, as shown in table \ref{tab:opt-rsvg-sota}. Figure \ref{fig:inference} shows the inference results of the adapted Grounding DINO on both datasets.

Table \ref{SOTA comparison} highlights the effectiveness of adapter-based fine-tuning to adapt OFA to visual grounding tasks. While BitFit—updating only 0.038\% of parameters—achieved a low mean IoU of 37.7\% due to its limited capacity to adapt to task-specific features, adapters reached 62.23\% mean IoU with 99.23 \% parameter efficiency, striking a strong balance between performance and computational cost. Moreover, results on the OPT-RSVG test set (Table \ref{tab:opt-rsvg-sota}) show that, although the adapter-tuned OFA model (Pr@0.5 = 66.38\%, mean IoU = 41.67\%) did not surpass LPVA, it remains competitive, especially given its efficiency. These findings emphasize that while PEFT methods, such as adapters, may not always match SOTA accuracy, they offer practical value for resource-constrained deployments.

\section{Conclusion}

The results collectively emphasize the effectiveness of PEFT techniques in adapting both specialized and foundation models. For Grounding DINO, LoRA demonstrates that minimal parameter updates can surpass SOTA performance, while OFA’s use of adapters highlights the flexibility of foundation models for specialized domains, in contrast to BitFit’s limited capacity. By limiting the trainable parameters, PEFT techniques provide an efficient, low-cost solution for domain-specific tasks while preserving accuracy. Future work may explore hybrid PEFT methods and extend adaptation to other remote sensing vision-language tasks.


\small
\bibliographystyle{IEEEtranN}
\bibliography{references}

\end{document}